\documentclass[journal]{IEEEtran}
\usepackage{cite}

\ifCLASSINFOpdf
  \usepackage[pdftex]{graphicx}
  \usepackage{subcaption} 
\else
\fi
\usepackage{amsmath}
\usepackage{amssymb}
\usepackage{booktabs}
\usepackage{balance}

\usepackage{placeins}

\usepackage{url}

\begin{document}
%
\title{Real-Time Thermal-Inertial Odometry on Embedded Hardware for High-Speed GPS-Denied Flight}
%
%
%

\author{Austin~Stone,
        Mark~Petersen,
        and~Cammy~Peterson
\thanks{A. Stone and C. Peterson are with Brigham Young University, Provo, UT 84602 USA.}}

%
%

\markboth{Preprint, 2026}%
{Stone \MakeLowercase{\textit{et al.}}: Real-Time Thermal-Inertial Odometry for High-Speed GPS-Denied Flight}
%



\maketitle

\begin{abstract}
We present a real-time monocular thermal–inertial odometry system designed for high-velocity, GPS-denied flight on embedded hardware. The system fuses measurements from a FLIR Boson+ 640 longwave infrared camera, a high-rate IMU, a laser range finder, a barometer, and a magnetometer within a fixed-lag factor graph. To sustain reliable feature tracks under motion blur, low contrast, and rapid viewpoint changes, we employ a lightweight thermal-optimized front-end with multi-stage feature filtering. Laser range finder measurements provide per-feature depth priors that stabilize scale during weakly observable motion. High-rate inertial data is first pre-filtered using a Chebyshev Type~II infinite impulse response (IIR) filter and then preintegrated, improving robustness to airframe vibrations during aggressive maneuvers. To address barometric altitude errors induced at high airspeeds, we train an uncertainty-aware gated recurrent unit (GRU) network that models the temporal dynamics of static pressure distortion, outperforming polynomial and multi-layer perceptron (MLP) baselines. Integrated on an NVIDIA Jetson Xavier NX, the complete system supports closed-loop quadrotor flight at 30~m/s with drift under 2\% over kilometer-scale trajectories. These contributions expand the operational envelope of thermal–inertial navigation, enabling reliable high-speed flight in visually degraded and GPS-denied environments.
\end{abstract}


%
\IEEEpeerreviewmaketitle



\section{Introduction}
\label{sec:intro}
Accurate state estimation in GPS-denied and visually degraded environments remains a core challenge in the field of aerial robotics. Visually degraded environments, characterized by poor illumination or obscurants such as smoke or fog, are encountered in a wide range of real-world applications, including military operations, nighttime urban navigation, infrastructure inspection, subterranean exploration, and search and rescue missions. In many of these scenarios, rapid maneuverability is also a critical requirement, placing further demands on the robustness and responsiveness of the estimation system. In this work, we address the intersection of these challenges by presenting a monocular thermal-inertial odometry (TIO) system explicitly designed for high-velocity, GPS-denied flight on an embedded platform.

To address the challenges posed by GPS-denied environments, visual-inertial odometry (VIO) has emerged as a leading solution. VIO systems leverage the complementary characteristics of cameras and inertial measurement units (IMUs) \cite{msckf}. In general, cameras offer low drift observations of structure but lack metric scale, while IMUs provide high-rate metric motion estimates, albeit subject to integration and bias drift. We augment this core VIO framework with three additional low-cost sensors: a laser range finder (LRF), a barometer, and a magnetometer. We incorporate their measurements by adding corresponding factors to the factor graph in a VINS-Fusion–style formulation \cite{vins-fusion}. The LRF provides direct, metric depth information that aids in both initializing feature depths in the monocular pipeline and maintaining scale consistency during constant acceleration \cite{xVio}. The barometer--a standard component on most quadcopters--offers high-rate pseudoaltitude measurements, though it is prone to error in high-velocity maneuvers due to pressure disturbances \cite{nn-baro}. Lastly, the magnetometer constrains yaw drift. While both roll and pitch are observable through the IMU, yaw is not \cite{vins-mono}. Yaw drift is a significant source of error in VIO, but it can be mitigated even with a relatively low-quality magnetometer. 

Although traditional VIO algorithms have proven extraordinarily successful under moderate speeds and good lighting, several factors limit their applicability in high-speed, low-illumination, or obscurant-rich regimes. First, standard RGB cameras suffer from low signal-to-noise ratios in low light, and narrowing exposure to mitigate motion blur often yields further underexposed images. Second, RGB cameras are unable to penetrate environmental obscurants, such as smoke or fog. However, longwave infrared (LWIR) thermal cameras offer a compelling alternative for visually degraded conditions, as they capture radiometric contrasts that persist irrespective of ambient illumination. They can also penetrate through obscurants. When integrated into a visual-inertial framework, this yields thermal-inertial odometry, which combines the robustness of inertial sensing with the visibility advantages of LWIR imagery \cite{Khattak}.

While LWIR cameras offer key advantages, they are not a panacea. They can struggle in scenes dominated by uniform temperatures—such as snow, rain, or recently wet surfaces—where thermal contrast diminishes and reflections or cooling effects distort radiometric cues. In these situations, an electro-optical (EO) camera may outperform thermal vision. Thermal images also typically exhibit lower contrast than visible spectra, since most natural scenes yield smooth temperature gradients rather than sharply textured features. Unlike EO cameras whose photodiodes' charge is fully dumped in microseconds, LWIR camera thermistors have a time constant on the order of several milliseconds \cite{flir2019specs}. The residual heat in the thermistor means that some of the previous scene still contributes to the current scene, causing additional blurring. Additionally, LWIR cameras suffer from significant fixed pattern noise (FPN) arising from temperature-varying, pixel-to-pixel nonuniformities in the detector array. Although in-field non-uniformity correction (NUC) routines can attenuate FPN, they introduce intermittent ``dropout'' periods--often ranging from 250 ms to 1s--during which the sensor ceases to output valid frames \cite{flir2019specs}. In high-speed flight such dropouts can cause significant drift in the state estimate as features disappear and IMU biases are not corrected properly. Although the mitigation provided by NUCs is normally effective in removing the majority of the FPN, artifacts and low-frequency noise can still remain. This is especially true when operating in the cold.

Additional challenges are introduced by the high-velocity, low-altitude flight regime targeted by our system. Unlike high-altitude aircraft, which can employ terrain relative navigation (TRN) techniques that match radar or optical profiles to stored digital terrain maps for drift correction, low-altitude operation precludes such approaches—the field of view (FOV) is too narrow and the terrain changes too rapidly relative to onboard map resolution \cite{Jarraya2025}. Instead, navigation must rely on real-time visual-inertial estimation using locally observed structure, which must remain stable even when operating close to the ground at high speed. While prior work such as S-MSCKF \cite{SMSCKF} has demonstrated stereo RGB VIO at speeds up to 17.5 m/s, our system operates at 30 m/s using monocular thermal imagery. At such velocities, new issues manifest and existing struggles are exacerbated. Motion blur becomes more pronounced, compromising visual feature tracking. Feature tracks shorten due to the reduced time features remain in view, increasing data association challenges and decreasing the impact a feature can have on state estimation. Processing latency becomes more detrimental as aggressive maneuvers demand low-latency state estimates to maintain control stability. The onboard embedded processor must therefore perform robust estimation in real time under strict computational constraints. Furthermore, high-speed flight exacerbates mechanical vibrations, impairing IMU measurement quality. 

Another challenge of high-speed flight is maintaining an accurate altitude estimate. While VIO provides reliable short-term vertical motion estimates, its altitude inevitably drifts over longer trajectories, making global measurements important for sustained flight. Barometric pressure measurements typically play this role and are used to stabilize the vertical state in VIO systems. However, barometers are highly sensitive to aerodynamic disturbances: as the vehicle’s airspeed increases, static underpressure at the barometer’s port causes the reported altitude to rise artificially \cite{nn-baro}. 
In our experiments, these high-speed disturbances produced altitude errors exceeding 15~m, large enough to decrease estimator performance and pose a serious risk during low-altitude flight. To the best of our knowledge, no prior work has investigated how static pressure position error (SPPE) impacts VIO-based state estimation.

This paper addresses key challenges in visual–inertial odometry (VIO) for high-speed flight using a thermal camera.  We introduce a unified TIO estimator that tackles (1) high-speed flight, (2) monocular thermal imaging, and (3) dynamic barometer compensation, all while meeting the strict compute and power constraints of embedded hardware. By addressing the intersecting challenges of high-speed flight, thermal imaging, and dynamic barometric disturbances, we expand the operational envelope of autonomous quadrotors, enabling reliable navigation in previously inaccessible environments.

Our main contributions are as follows:
\begin{itemize}
    \item We introduce an efficient and robust feature tracking and filtering pipeline tailored to the unique challenges of high-speed thermal imaging, including motion blur, low contrast, and FPN.
    \item We propose a novel barometric compensation algorithm that models the temporal dynamics of barometric error and achieves superior performance compared to existing techniques.
    \item We propose an IMU prefiltering step that improves estimator stability during aggressive maneuvers.
    \item We present the first monocular thermal-inertial odometry system demonstrated to support closed-loop flight at velocities reaching 30~m/s. Our system is validated through rigorous hardware experimentation.
\end{itemize}

The remainder of this paper is organized as follows. Section \ref{sec:background} reviews related work in high-speed VIO, thermal-inertial odometry, and barometric compensation. Section \ref{sec:setup} describes our system setup, including sensor synchronization, calibration routines, and NUC scheduling. Section \ref{sec:methods} presents the algorithmic design, from front-end image enhancement and feature tracking to back-end smoothing. Section \ref{sec:methods} also details our barometer compensation model and its integration into the smoothing framework. Section \ref{sec:results} reports extensive experimental results comparing feature detection accuracy, IMU preprocessing effectiveness, barometer error reduction, trajectory accuracy, and computational performance across multiple flight velocities and environmental conditions. Finally, Section \ref{sec:conclusion} concludes and outlines avenues for future research.


\section{Background}
\label{sec:background}

This work lies at the intersection of two primary domains—TIO and barometric compensation—while building on foundational insights from VIO. Although TIO inherits its algorithmic structure from VIO frameworks, the sensing modality fundamentally changes the problem: thermal imagery behaves differently from visible-light imagery, requiring dedicated front-end processing, tracking, and robustness strategies for low-texture or noisy scenes. While each of these areas has been explored independently, their integration in the context of high-speed, low-visibility aerial navigation remains under-studied. This section reviews the state of the art across each domain, highlighting both their evolution and limitations.

\subsection{VIO}
\label{sec:background:VIO}

VIO has evolved as a primary solution for state estimation in GPS-denied environments, relying on the complementary strengths of cameras and IMUs. Tightly coupled methods dominate the field and can be divided into two broad categories: filter-based and optimization-based approaches.


Filter-based methods were the earliest practical implementations of VIO, leveraging variants of the Kalman filter to fuse high-rate inertial data with lower-rate camera observations. These approaches typically employ either an Extended Kalman Filter (EKF) or an Unscented Kalman Filter. In both cases, the IMU is used to propagate the state forward in time, while camera-derived feature tracks provide constraints on IMU drift.

A seminal system in this category is the Multi-State Constraint Kalman Filter (MSCKF)~\cite{msckf}. The MSCKF introduces the concept of augmenting the filter state with a short history of camera poses. When a feature is observed across multiple frames, its measurements are projected onto these poses to generate multi-view constraints. These constraints are formed once the feature track terminates and are then used as delayed measurement updates to the filter. Consequently, the visual information cannot be used immediately. By marginalizing feature states once their constraints are incorporated, the MSCKF prevents unbounded state growth while still exploiting multi-frame geometry, enabling real-time operation on embedded processors. Numerous derivatives have refined this architecture, such as S-MSCKF~\cite{SMSCKF} for stereo cameras and robocentric formulations (R-VIO)~\cite{rvio} that reduce sensitivity to linearization drift.

Another notable contribution is ROVIO~\cite{rovio}, which integrates direct photometric error into the EKF update step, effectively coupling pixel intensity alignment with inertial propagation. This innovation demonstrated that filter-based frameworks can accommodate not only geometric residuals but also direct image alignment, improving accuracy in low-texture scenes. Similarly, loosely coupled formulations have been proposed that fuse outputs from existing visual odometry pipelines (e.g., monocular SfM) with inertial estimates via an EKF, offering modularity and ease of implementation \cite{loosely-coupled-1, loosely-coupled-2}.

Filter-based VIO methods offer several practical advantages in embedded settings. Their update structure produces state estimates with very low latency, which is essential for fast control loops on agile aerial robots. Because the state dimension remains small and updates are local, these methods are computationally lightweight and well-suited for processors with limited resources. Moreover, their modular formulation allows additional sensors such as magnetometers, barometers, GPS pseudoranges, or rangefinders to be incorporated with minimal changes to the underlying filter equations.

Despite these advantages, limitations arise from the reliance on linearization about a single operating point \cite{vins-fusion}. Over long trajectories, small errors in the linearization accumulate, leading to inconsistency and drift. Filters also struggle to handle loop closures or delayed updates gracefully, since past states are no longer available once marginalized. As a result, while filter-based VIO remains popular in embedded and time-critical applications, their accuracy is typically outperformed by modern optimization-based approaches, which revisit and jointly refine past states. Nevertheless, the ideas pioneered by MSCKF \cite{msckf} and its descendants remain influential: sliding-window marginalization, multi-frame feature constraints, and lightweight sensor fusion strategies continue to inform current-generation systems.


With growing onboard compute, optimization-based methods emerged as the dominant paradigm for VIO. These systems formulate estimation as a nonlinear least-squares problem over a sliding window of recent states, retaining multiple poses, velocities, and biases in the active set \cite{vins-mono}. Measurements are incorporated as residual factors, and the joint optimization refines all variables simultaneously. By revisiting past states, optimization reduces linearization error, improves robustness to outliers, and naturally accommodates delayed updates.

OKVIS~\cite{okvis} demonstrated the viability of tightly coupled optimization of visual reprojections and inertial constraints in real time, showing markedly lower drift than EKF-based methods. VINS-Mono~\cite{vins-mono} extended this approach to monocular cameras with loop closure and pose graph optimization for global consistency, while VINS-Fusion~\cite{vins-fusion} generalized the factor graph to support stereo vision and additional sensors, illustrating the modularity of the framework. 

A key enabler for these systems is IMU preintegration. The idea of summarizing high-rate inertial data between camera frames into a single compound measurement dates back at least to Lupton and Sukkarieh~\cite{preintegration2}, who introduced a preintegrated inertial measurement model for graph-based smoothing. Forster et al.~\cite{preintegrationForster} refined this concept by deriving closed-form expressions on the $\mathrm{SO}(3)$ manifold, providing both accuracy and efficiency while avoiding repeated reintegration of raw IMU data. 

Recent work has focused on scalability and efficiency. ICE-BA~\cite{ICEBA} introduced incremental relinearization, updating only Jacobians affected by large state changes, thereby achieving low-latency operation even with extended windows. XR-VIO~\cite{xr-vio} highlighted the importance of robust initialization and hybrid feature matching, demonstrating that front-end reliability is inseparable from back-end optimization accuracy.

The advantages of optimization-based approaches include reduced drift through joint refinement, flexible integration of heterogeneous sensor modalities, and delayed consistency by incorporating constraints once sufficient parallax is available. However, these benefits come at a computational cost: solving a nonlinear least-squares problem at every step is demanding, particularly on embedded platforms where processing budgets are limited.

To maintain real-time performance, practical systems must make compromises. Many optimization-based VIO frameworks restrict the number of iterations per update cycle, meaning the solver may not fully converge at each timestep. Furthermore, aggressive marginalization is required to bound computational cost, which reintroduces some of the linearization errors that optimization frameworks aim to avoid. These issues result in increased latency and drift over long trajectories.

Our system builds directly on the optimization-based tradition exemplified by VINS-Fusion \cite{vins-fusion} and ICE-BA \cite{ICEBA}. We adopt a fixed-lag smoothing formulation with joint optimization of poses, velocities, and biases, employ the manifold-based preintegration model of Forster et al. \cite{preintegrationForster}, and use ICE-BA–style incremental relinearization to sustain real-time performance. We further extend the factor graph with additional sensor modalities—IMU- and magnetometer-derived attitude, LRF depth priors, and barometric altitude—tailoring the optimization framework to the unique challenges of high-speed thermal-inertial navigation.

\subsection{Thermal-Inertial Odometry}
\label{sec:background:TIO}

\begin{figure}[t]
    \centering
    \begin{subfigure}[t]{0.48\linewidth}
        \centering
        \includegraphics[width=\linewidth]{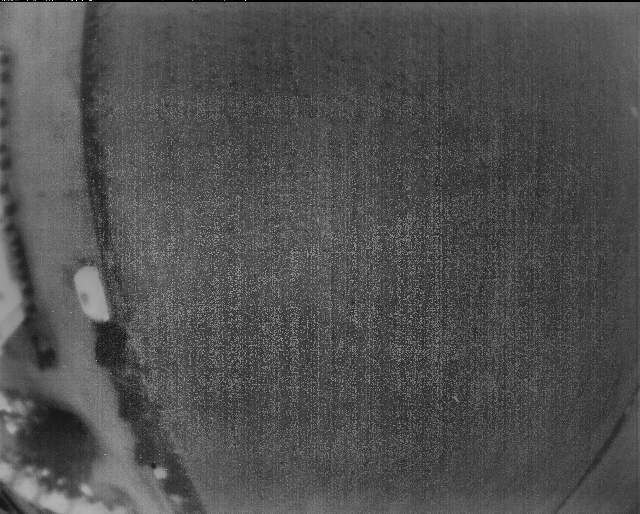}
        \caption{Degraded LWIR Image}
        \label{fig:bad_boson}
    \end{subfigure}
    \hfill
    \begin{subfigure}[t]{0.48\linewidth}
        \centering
        \includegraphics[width=\linewidth]{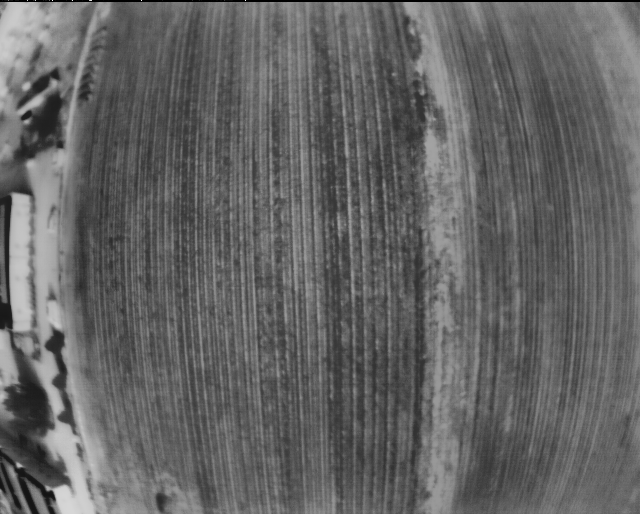}
        \caption{Good LWIR Image}
        \label{fig:good_boson}
    \end{subfigure}
    \caption{Comparison of degraded and good IR images. Degraded image shows characteristic cross-hatch pattern.}
    \label{fig:boson_comparison}
\end{figure}

Thermal-inertial odometry enables navigation in degraded environments (e.g., low-light, smoke) where RGB cameras fail. LWIR cameras are resilient to illumination and penetrate many obscurants, but TIO faces the following challenges:

\begin{enumerate}
    \item \textbf{Lower Spatial Resolution and Contrast:} LWIR sensors typically offer lower pixel resolutions and narrower fields of view compared to visible cameras (for example, the FLIR Boson+ provides a maximum horizontal FOV of approximately $95^\circ$). In addition, natural thermal scenes exhibit smooth radiometric gradients with limited texture, reducing both the number and reliability of trackable features.
    
    \item \textbf{FPN:} FPN results from nonuniform pixel response and is periodically compensated for by the sensor through a NUC routine, which involves comparing the sensor output against a uniform temperature source (often a shutter). See Figure \ref{fig:boson_comparison} for an example of FPN.

    \item \textbf{Image Dropout:} During NUC (typically lasting 250--1000~ms), the camera may freeze output or produce invalid frames, temporarily removing visual constraints from the estimator. This prevents correction of accumulating IMU bias drift or accelerometer noise during those intervals.
    
    \item \textbf{Thermal Lag:} LWIR cameras use microbolometers (thermal detectors that measure infrared radiation as heat), which have finite thermal time constants on the order of 5–15~ms. As a result, the output at time $t$ contains residual information from prior frames. During aggressive maneuvers or high vibration, thermal lag manifests as increased blur or smearing in the imagery, further degrading feature stability \cite{ThermalInertialFlighAtNight}.

\end{enumerate}









Recent TIO research spans a mix of learned and handcrafted approaches. Learned descriptor pipelines such as TP-TIO~\cite{tptio}, which adapts SuperPoint~\cite{superpoint} using synthetic thermal augmentation, improve feature robustness to LWIR artifacts. Other work focuses on selecting more informative regions: Khattak et al.~\cite{Khattak} use spatial entropy to guide direct photometric tracking in low-contrast scenes. Flow-based methods such as Self-TIO~\cite{self-tio} combine a KLT initializer with RAFT~\cite{raft} to achieve subpixel motion estimates, applying SVD-based denoising to mitigate FPN. End-to-end regressors like DeepTIO~\cite{deeptio} bypass feature extraction entirely by directly predicting pose from thermal imagery and IMU data, though they remain computationally demanding and less generalizable.

Handcrafted pipelines offer an alternative emphasizing efficiency and interpretability. Systems such as ETIO~\cite{edge-based-tio} employ edge detection, binarization, and KLT tracking, while related methods adopt direct alignment or distance-transform objectives inspired by analogous VIO literature~\cite{edge-vio-1, edge-vio-2}. These methods continue to be the most tractable for embedded deployment due to their low latency and modest compute requirements. 

Although deep learning–based methods have demonstrated promising performance gains with thermal imagery, their inference times remain incompatible with the strict latency and power constraints of onboard drone hardware. For example, Self-TIO requires over 30 ms per frame on a laptop-grade GPU and TP-TIO takes 30 ms just to track features. As a result, handcrafted pipelines remain the most practical solution for embedded TIO in fast, closed-loop flight. Our system follows this philosophy: inspired by XR-VIO~\cite{xr-vio}, we combine contrast limited adaptive histogram enhancement (CLAHE), gradient-based filtering, and descriptor validation to maintain stable tracks through motion blur and FPN. This hybrid design preserves the computational efficiency needed for real-time high-speed navigation while directly addressing the visual challenges of LWIR imagery.

\subsection{High-Speed Navigation}
High-speed navigation has been extensively explored in high-altitude platforms using TRN. Systems such as TERCOM\cite{TERCOM1, TERCOM2} and SITAN \cite{SITAN} fuse radar altimetry with digital terrain maps to provide drift-free state estimates for fixed-wing aircraft and missiles operating at high velocities. More recently, vision-based TRN approaches have demonstrated impressive results. Maggio et al.~\cite{maggio2023visionbasedterrainrelativenavigation} report sub-55~m average error on a suborbital Blue Origin flight using downward-facing camera imagery and inertial sensing to match observed terrain to satellite maps at speeds exceeding 240~m/s. These methods work best at high altitude, where broad terrain features are visible and can be matched to known maps. 

In contrast, low-altitude navigation presents different challenges—parallax is stronger but the visible area is limited. Within this regime, VIO is more commonly used \cite{Jarraya2025}. However, existing VIO systems degrade at high speed due to motion blur, shortened feature tracks, and increased control loop delays. Sun et al.~\cite{SMSCKF} demonstrated that S-MSCKF can support stereo RGB-based VIO up to 17.5~m/s, but it relies on stereo vision and assumes favorable lighting. Most public datasets omit thermal imagery and operate at modest speeds—for example, the EuRoC Micro Air Vehicle dataset rarely exceeds 5~m/s~\cite{Euroc}, and SwiftBase~\cite{swiftbase} provides moderate-speed stereo monochrome imagery at 200~Hz. Racing drone datasets reach up to 28~m/s~\cite{kaufmann2023champion}, but typically involve RGB cameras in structured indoor or short outdoor tracks. 

To the best of our knowledge, our work is the first to demonstrate real-time, monocular thermal–inertial odometry at 30~m/s in low-altitude flight on embedded hardware with closed-loop control.

\subsection{Barometric Altitude Compensation}
\label{sec:background:baro}

Barometric sensors provide altitude estimates by measuring atmospheric pressure. Although attractive for their low cost and simplicity, they are susceptible to errors induced by high-speed motion. This SPPE arises when airflow over the barometer’s port introduces local underpressure, leading to spurious fluctuations in measured altitude. This effect increases with vehicle velocity and can severely degrade estimator performance during aggressive maneuvers.

Existing compensation strategies are heuristic. For example, PX4 \cite{px4} and ArduPilot \cite{ardupilot} recommend polynomial fitting based on manually collected calibration data. A more recent approach uses a multi-layer perceptron (MLP) neural network to learn a mapping from velocity to barometric correction \cite{nn-baro}. However, such feedforward models disregard the temporal structure of SPPE.

In contrast, our system employs a gated recurrent unit (GRU) network \cite{gru} to model the temporal dependencies of barometric error. The learned correction is then integrated into the factor graph as an altitude constraint, enabling corrections that improve estimation and long-term drift. To our knowledge, this is the first demonstration of recurrent barometric correction applied within a visual-inertial framework.


\section{Setup}
\label{sec:setup}

Reproducibility in real-time state estimation for aerial robotics requires meticulous attention to hardware synchronization and system configuration. In this section, we detail the setup procedures and calibration protocols that underpin our experimental platform.

\subsection{Hardware Configuration}
\label{sec:setup:hardware}

All estimation processes are executed on an NVIDIA Jetson Xavier NX. Given the presence of a non-real-time operating system, precise timestamping of sensor data is nontrivial. To address this, we employ a dedicated micro-controller, which we will call the sensor manager (SM) for time-critical sensor handling and synchronization.

The IMU is hardware-triggered at 1200~Hz. Raw data is filtered onboard using the infinite impulse response (IIR) filter pipeline described in Section~\ref{sec:methods:imupreprocessing}, and then downsampled to 120~Hz before being forwarded to the Xavier NX. The thermal camera (FLIR Boson+ 640) operates at 30~Hz and is hardware-triggered by the SM. Each camera trigger is time-aligned with a corresponding IMU trigger, and the associated IMU sample is tagged as image-synchronized. The images are read in by the Xavier NX and timestamped with the image-aligned IMU measurement timestamp.

To maintain a consistent time base between the SM and the Xavier NX, we employ a hardware-assisted synchronization scheme. Each second the SM outputs a synchronization pulse and sends a corresponding timestamped message that is captured by the NX and used to estimate clock offset and drift. The offset is applied to translate all image and IMU timestamps into the unified SM time frame. This ensures sub-millisecond temporal consistency across sensors while avoiding the latency and jitter typically associated with software-only synchronization.

Additional sensors are time-stamped by the SM and forwarded asynchronously to the Xavier NX. To provide metric depth information, the LRF is rigidly mounted and bore-aligned with the camera, such that its measurement represents the depth to the center of the image.

An instance of PX4 runs on a separate microcontroller and serves as the flight controller on our vehicle. For closed-loop flight, it has access to its own dedicated IMU and receives positional information from the TIO pipeline running on the Xavier NX as it becomes available (30 Hz).

\subsection{Camera Calibration}
\label{sec:setup:cameracalibration}

Thermal-inertial calibration is performed using Kalibr \cite{kalibr} and a custom calibration target designed for long-wave infrared (LWIR) imaging. The target comprises a reflective board with absorptive materials tracing an array of ArUco markers \cite{aruco}. The target is placed flat on the ground so as to reflect the cold sky. A standard calibration motion sequence is performed to excite the necessary modes for observability. However, occlusion of the sky during the calibration motion sequence somewhat limits calibration quality. We adhered to Kalibr's recommended sensor rates for calibration: 20~Hz for the camera and 200~Hz for the IMU.

\subsection{Flat-Field Correction Scheduling}
\label{sec:setup:FFC}
The Boson+ thermal camera requires periodic flat-field corrections (FFCs) to compensate for internal temperature drift and non-uniformity artifacts. This creates a dropout window where no features are available for tracking. Improperly timed FFCs can therefore exacerbate estimator drift by disrupting feature continuity. To mitigate this, we implement a dynamic FFC scheduler with the following policy:
\begin{itemize}
    \item The Boson+ requests an FFC based on internal metrics.
    \item If the current linear velocity is below 5~m/s and angular velocity is below 0.5~rad/s, the FFC is triggered immediately if requested by the Boson+.
    \item Otherwise, the FFC is deferred for up to 2 minutes, or until motion drops below thresholds—whichever occurs first.
    \item An initial FFC is always triggered prior to takeoff to ensure consistent image quality.
\end{itemize}

This strategy balances the need for thermal correction with the requirement for uninterrupted visual-inertial tracking, particularly during high-speed flight segments.


\section{Methods}
\label{sec:methods}

In this section, we outline our overall algorithmic architecture, and then detail the individual components that collectively ensure robust state estimation under the demanding conditions outlined previously.

\subsection{Algorithm Overview}

\begin{figure*}[ht]
    \centering
    \includegraphics[width=\textwidth]{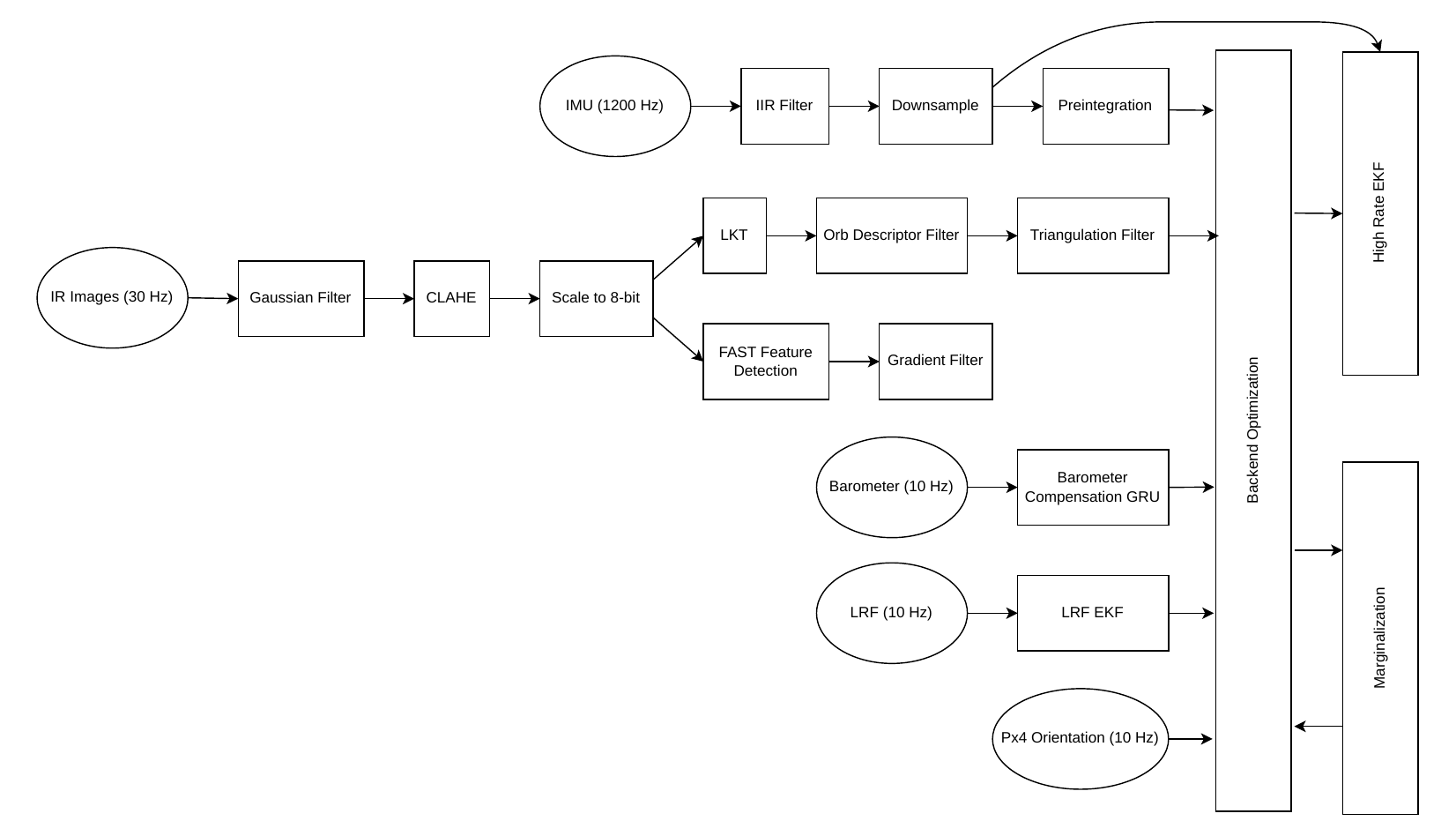}
    \caption{Overview of our monocular thermal-inertial odometry pipeline.}
    \label{fig:algorithm_overview}
\end{figure*}

Figure \ref{fig:algorithm_overview} illustrates our complete estimation pipeline. Raw thermal imagery from the FLIR Boson+ camera first undergoes preprocessing via Gaussian filtering and CLAHE \cite{CLAHE} to enhance image contrast and reduce FPN. Features are then detected using a FAST corner detector \cite{fast-features}. Optical flow tracking is employed to maintain feature tracks across consecutive frames, supplemented by essential matrix outlier rejection \cite{seven-point}. Feature tracking robustness is enhanced through gradient-based filtering and ORB descriptor matching \cite{orb}.

Simultaneously, high-rate IMU measurements (1200Hz) undergo preprocessing through a Chebyshev Type II IIR filter \cite{scipy} to attenuate mechanical vibrations, followed by downsampling to 120Hz. The filtered and downsampled IMU data is preintegrated\cite{preintegrationForster}, providing inertial constraints between successive optimization windows.

Feature tracks are subsequently validated through triangulation. Validated features and IMU preintegration results feed into our Levenberg-Marquardt Preconditioned Conjugate Gradient (LMPCG) backend optimization \cite{ICEBA, pcg, BundleAdjustmentInTheLarge, ConjugateGradientBundleAdjustment, PushingTheEnvolopeOfModernBundleAdjustment}, formulated as a sliding-window nonlinear least squares optimization that jointly estimates navigation states (pose, velocity, biases) and feature inverse depths.

To address altitude errors introduced by SPPE on barometric measurements during high-speed maneuvers, we employ an uncertainty-aware GRU neural network. The GRU dynamically corrects the barometric altitude measurements by capturing temporal dependencies, and these corrected measurements serve as additional altitude constraints in the optimization framework.

Finally, the resulting optimized state estimates are fused into a lightweight high-rate EKF to provide low-latency pose updates for closed-loop control. The EKF is propagated by the IMU and is updated by the backend optimizer, enabling fast, drift-corrected estimates suitable for aggressive flight. This structured approach, leveraging carefully tailored preprocessing, robust feature filtering, and novel barometric compensation, significantly improves estimator accuracy and reliability under demanding flight conditions.

\subsection{Initialization}
\label{sec:methods:initialization}

To ensure safe deployment and consistent state estimation, we assume the vehicle remains stationary during the estimator initialization phase—a reasonable assumption given the use of launch platforms in many aerial robotic applications. Under this assumption, we initialize the local reference frame by fixing the initial position to the origin and aligning orientation with gravity and magnetic north.

We seek to estimate five quantities: gyroscope bias, accelerometer bias, initial roll, initial pitch, and initial yaw. Given the stationary condition, we leverage statistical properties of high-rate sensor data to recover these values.

\paragraph*{Gyroscope Bias} 
The gyroscope bias $\mathbf{b}_g$ is estimated as the average of 500 gyroscope measurements $\{\boldsymbol{\omega}_i\}$, exploiting the fact that true rotational velocity should be zero.

\paragraph*{Roll, Pitch, and Accelerometer Bias} 
We formulate a nonlinear optimization to jointly estimate the accelerometer bias $\mathbf{b}_a$, roll $\phi$, and pitch $\theta$. Let $\mathbf{a}$ denote the average of 500 accelerometer measurements and $\mathbf{g} = [0, 0, g]^\top$ denote the gravity vector in the inertial frame. The error in our model can then be expressed as:
\[
\mathbf{e} = \mathbf{a} - \mathbf{b}_a + \mathbf{R}_{\text{pitch}}(\theta) \mathbf{R}_{\text{roll}}(\phi) \mathbf{g}
\]
We can further define a weighted objective as:
\[
J(\mathbf{x}) = \|\mathbf{e}(\mathbf{x})\|^2 + w \|\mathbf{b}_a\|^2,
\]
where $\mathbf{x} = [\mathbf{b}_a^\top, \phi, \theta]^\top$ and $w$ is a configurable weight. The first term enforces alignment between the measured acceleration and the gravity vector based on the estimated orientation and bias. However, because roll, pitch, and accelerometer biases are not fully observable from accelerometer data alone in a stationary scenario, the problem is underconstrained. To address this, we include the second term as a regularization penalty that minimizes the magnitude of the accelerometer bias $\mathbf{b}_a$. We compute the Jacobian $\mathbf{J}$ and Hessian $\mathbf{H}$ of this objective and solve the resulting nonlinear system using a Truncated Newton Trust Region solver.

The Jacobian $\mathbf{J} \in \mathbb{R}^{3 \times 5}$ of the error function with respect to the state $\mathbf{x}$ is:
\[
\mathbf{J} =
\begin{bmatrix}
-1 & 0 & 0 & 0 & -c_\theta g \\
0 & -1 & 0 & c_\phi c_\theta g & -s_\phi s_\theta g \\
0 & 0 & -1 & -s_\phi c_\theta g & -c_\phi s_\theta g
\end{bmatrix},
\]
where $s_\phi = \sin(\phi)$, $c_\phi = \cos(\phi)$.

The Hessian $\mathbf{H} \in \mathbb{R}^{5 \times 5}$ is given by:
\[
\mathbf{H} = 2w \mathbf{J}^\top \mathbf{J} + 2
\begin{bmatrix}
\mathbf{I}_{3 \times 3} & \mathbf{0} \\
\mathbf{0} & \mathbf{0}
\end{bmatrix}
+ 2w
\left(
\frac{\partial \mathbf{J}}{\partial \phi}^\top \mathbf{e} +
\frac{\partial \mathbf{J}}{\partial \theta}^\top \mathbf{e}
\right)
\]

\paragraph*{Yaw Initialization}
Yaw $\psi$ is unobservable from accelerometers and gyroscopes alone. To resolve global orientation, we compute the average of fourteen orientation measurements from the PX4 (the same time interval as the 500 IMU measurements) and align the yaw component with magnetic north.

The above procedure provides a reliable initialization for state estimation, aligning the inertial frame with Earth’s gravity and magnetic north, while compensating for sensor biases.

When the vehicle is stationary on the ground, there may be no reliable visual features available to constrain position. To prevent estimator drift in this underconstrained setting, we impose a soft prior on states as they are added, anchoring them to the origin. This prevents the estimator from slowly drifting before takeoff.

\subsection{IMU Preprocessing}
\label{sec:methods:imupreprocessing}

Our system samples the onboard IMU at 1200~Hz, but due to bandwidth constraints in the communication pipeline, only downsampled data at 120~Hz is forwarded to the front end for estimation. Although the IMU is mounted on a vibration-isolated platform, residual mechanical noise can corrupt accelerometer and gyroscope readings. To mitigate high-frequency noise introduced by airframe vibrations and the resultant aliasing when downsampling, we first apply an IIR filter to the raw IMU data. 

We utilize a first-order Chebyshev Type II IIR filter with a 50~Hz cutoff frequency. The filter coefficients are: $A = [1, -0.97104422]$ and $B = [0.01447789, 0.01447789]$. Filter coefficients were obtained using the scipy signal package \cite{scipy}. This filter is applied independently to each axis of the accelerometer and gyroscope. As demonstrated in Section~\ref{sec:results:imuprefiltering}, this filtering step significantly enhances preintegration performance at high speeds.

Following filtering and downsampling, we perform IMU preintegration using the 120~Hz stream following the method outlined by Forster et al.~\cite{preintegrationForster}.

\subsection{Zero-Velocity Detection and Feature Consistency}
\label{sec:methods:zerovelocity}

To further improve the observability of velocity and IMU biases, we employ a zero velocity detection module that continuously monitors IMU and optical flow data, similar to OpenVINS \cite{OpenVins}. The detector computes sliding-window statistics of accelerometer and gyroscope measurements, as well as gravity-compensated acceleration values. When the mean and standard deviation of these quantities fall within predefined thresholds, the platform is classified as being in a constant velocity state. If constant velocity is detected, an attitude factor is added in the same way as described below, but using gravity as the reference vector and the observed acceleration as the measurement. Zero velocity is inferred when constant velocity coincides with optical flow magnitude and covariance near zero. Upon detection, a zero-velocity factor is added to the factor graph to reduce drift during stationary intervals.

\subsection{Image Preprocessing}
\label{sec:methods:imagepreprocessing}

We acquire imagery from the pre-automatic gain control (pre-AGC) 16-bit digital video output of the FLIR Boson+ thermal camera. This configuration minimizes processing latency and provides access to the full dynamic range of the raw thermal signal prior to onboard rescaling. 

Although our feature tracking pipeline relies on 8-bit image processing routines from OpenCV, we first perform preprocessing in the native 16-bit domain to preserve as much information as possible. The preprocessing pipeline consists of three primary stages:

\begin{enumerate}
    \item \textbf{FPN Suppression:} A Gaussian filter with a $3\times3$ kernel and $\sigma=1$ is applied to the raw 16-bit image. This attenuates high-frequency FPN while preserving coarse thermal structures.

    \item \textbf{Contrast Enhancement:} We apply CLAHE \cite{CLAHE} directly on the smoothed 16-bit image. CLAHE locally enhances radiometric contrast, improving feature detectability in regions with weak thermal gradients. However, during scenes with rapid changes in thermal intensity, CLAHE accentuates the disparity, which can affect feature stability.

    \item \textbf{Quantization:} The contrast-enhanced image is then scaled and converted to an 8-bit grayscale image for use with OpenCV feature detection and tracking algorithms.
\end{enumerate}

This staged approach preserves scene detail during enhancement while ensuring compatibility with standard computer vision tools. The use of pre-AGC imagery, in particular, avoids clipping artifacts and ensures that high-dynamic-range thermal variations are maintained throughout the processing pipeline.

\subsection{Feature Tracking}
\label{sec:methods:featuretracking}

Thermal features are detected using the FAST corner detector \cite{fast-features} applied to the preprocessed 8-bit images whenever the number of tracked features drops below a threshold. This will be referred to as the detection frame. To ensure spatial diversity and prevent clustering, detection is constrained by a binning mask that divides the image into a fixed grid. New features are only allowed in bins that are currently unoccupied, thereby enforcing a minimum spacing criterion. If multiple features are added to a bin during detection, only the feature with the highest detection score is kept. 

Following detection, each candidate feature undergoes a gradient-based quality assessment akin to that used in the Harris detector \cite{Harris1988}. This filter computes the gradient magnitude at each feature location in the detection frame using a 5$\times$5 Sobel kernel \cite{sobel} in both horizontal and vertical directions. A feature is retained if its gradient magnitude exceeds a predefined threshold, reflecting the expectation that salient features in thermal imagery typically reside on edges or corners. This heuristic leverages the structural coherence of edges and corners to improve feature reliability.

Once accepted, features are tracked between frames using the KLT optical flow method \cite{lkt}. Each feature is associated with an ORB descriptor \cite{orb} computed at the time of detection in the detection frame. During tracking, the new feature location is validated by comparing the stored descriptor against the descriptor at the new location in the current frame. If the descriptor mismatch exceeds a threshold, the feature is marked as not found in the current frame. By just using the ORB descriptor as a check, we eliminate the time-intensive combinatorial problem of comparing all descriptors. Any feature that is not successfully tracked between frames is pruned from the set. Features are also pruned if more than a specified number are tracked into the same bin.

To further suppress erroneous matches, a 7-point RANSAC check \cite{seven-point}\cite{RANSAC} utilizing the essential matrix is employed to identify outliers relative to the dominant scene motion. This validation step is conducted between the last detection frame and the current frame to leverage greater parallax. If, after several frames, a feature is classified as an outlier in more than 50\% of recent frames, it is permanently removed. Observations that fail the essential matrix check are excluded from the back-end filter update.

To accommodate scene changes and maintain sufficient feature density, new features are periodically detected. Upon detection, each previously existing feature’s current descriptor is updated as the new reference descriptor for subsequent matching and its location is cached for use in the Essential Matrix check. Optical flow is seeded using the position of each feature in the previous frame to improve convergence and maintain temporal continuity. To maintain LRF-based scale cues, a new feature is detected in the central 20\% of the image whenever that region lacks active tracks.

\subsection{Feature Filtering}
\label{sec:methods:featurefiltering}

Due to the inherent challenges of tracking in thermal imagery, feature tracks can often be erroneous or unreliable even with the criteria mentioned above. To improve robustness, we incorporate a multi-stage filtering pipeline prior to backend inclusion, selecting only the most reliable feature measurements.

Candidate features are collected and evaluated in batches. For each candidate, a series of status checks are performed to assess geometric consistency. The resulting verified feature set is ranked according to a set of quality metrics (detailed below), and only the top-ranked subset—subject to a maximum new feature count—is passed to the backend at each update.

The primary filtering mechanism is a triangulation-based consistency check. Features are withheld from the backend until they accumulate at least ten valid observations or haven't received new observations in 0.5 seconds. At that point, 3D triangulation is attempted using a RANSAC-based algorithm modeled after OpenVINS~\cite{OpenVins, RANSAC}. This step is designed to reject spurious matches while preserving multi-frame geometric consistency. Triangulation always includes the first recorded measurement for the feature. Only the inlier subset from RANSAC is passed to the optimization backend. Once a feature has been accepted into the backend, subsequent measurements bypass this triangulation check to reduce computational cost and latency.

Several criteria are applied during triangulation and feature verification:
\begin{itemize}
    \item \textbf{Depth Sign Consistency:} If the estimated depth is negative, all associated measurements for that feature are discarded.
    
    \item \textbf{Depth Consistency with LRF:} Features near the center of the image are excluded if their triangulated depth differs significantly from the onboard LRF measurement. This cross-sensor consistency check improves depth reliability in low-parallax or ambiguous configurations.
    
    \item \textbf{Triangulation Quality Score:} To quantify the reliability of a triangulated 3D point, we compute a score based on the spatial arrangement of its viewing rays. Specifically, we assess how well-constrained the feature’s depth is by evaluating the condition number of the matrix that captures the geometry of the multi-view observations. A high score indicates that the viewing rays are nearly parallel, making depth estimation unreliable, while a low score corresponds to strong parallax and a well-conditioned triangulation. If the number of triangulated features exceeds a predefined threshold, only the top 80\% (by quality score) are passed to the backend.
    
    \item \textbf{Reprojection Error:} If any individual measurement within a track yields a factor error exceeding a defined threshold during factor graph construction, that measurement is not included in the backend. If too few valid measurements remain, the feature is excluded.
\end{itemize}

Triangulation is only performed when sufficient parallax is detected across the feature’s observations, ensuring depth observability. However, to accommodate transient features that may not persist long enough for full verification and the lack of parallax during hover, any feature that fails to meet the observation count or parallax criteria within the timeout window is still considered for backend inclusion if better alternatives are unavailable. This fallback guarantees timely inclusion of otherwise viable features during hover or aggressive maneuvering.

All thresholds and ranking weights (e.g., minimum observations, parallax thresholds, LRF agreement, etc.) were hand-tuned across more than 30 representative high-speed flight logs to balance estimation robustness with latency and throughput constraints.

This multi-tiered filtering stage significantly improves the reliability of feature tracks passed to the backend, particularly under high-speed and low-texture conditions typical of monocular thermal imagery.

\subsection{Feature Inclusion}
\label{sec:methods:featureinclusion}

Validated features are incorporated into the backend using a projection-based residual between two camera frames. Each feature is parameterized by a unit bearing vector $\mathbf{s}_i$ taken at the frame where the feature was first detected (the original root) and an inverse depth $\rho$ expressed in that frame. This inverse-depth formulation avoids over-parameterization and provides a compact representation for monocular measurements.

\paragraph*{Root-frame tracking through marginalization} Because the navigation state associated with the original root frame will be marginalized later, the system cannot assume that the feature's anchor frame persists in the factor graph. Instead of re-triangulating or redefining the feature when this happens, we maintain an accumulated SE(3) transform 
\[
{}^{\text{origRoot}}\mathbf{T}_{\text{currRoot}},
\]
where the current root is the oldest active keyframe with a feature observation. Whenever marginalization removes the previous root, this transform is updated. This strategy allows the feature to remain expressed in its original coordinate frame while shifting only the reference transformation used by the projection factor.

\paragraph*{Projection factor formulation}
Let ${}^{w}\mathbf{T}_{x}$ denote the body pose of navigation state $x$ in the world frame, and let ${}^{b}\mathbf{T}_{c}$ denote the camera-to-body extrinsic.  
If $x_i$ is the current root state and $x_j$ is the observation state, the relative transformation from the current root camera frame to the observation camera frame is
\[
{}^{c_i}\mathbf{T}_{c_j}
=
{}^{b}\mathbf{T}_{c}^{-1}
\,
{}^{w}\mathbf{T}_{x_j}^{-1}
\,
{}^{w}\mathbf{T}_{x_i}
\,
{}^{\text{currRoot}}\mathbf{T}_{\text{origRoot}}
\,
{}^{b}\mathbf{T}_{c}.
\]
Writing
\[
{}^{c_i}\mathbf{T}_{c_j} = ( {}^{c_i}\mathbf{R}_{c_j},\; {}^{c_i}\mathbf{t}_{c_j} ),
\]
the predicted 3D point in the observation frame is
\[
\mathbf{z}
= {}^{c_i}\mathbf{R}_{c_j}\, \mathbf{s}_i
  + \rho\, {}^{c_i}\mathbf{t}_{c_j}.
\]

To obtain a scale-invariant residual, we project $\mathbf{z}$ onto the plane orthogonal to the observed bearing $\mathbf{s}_j$:
\[
\mathbf{P} = \mathbf{I} - \mathbf{s}_j \mathbf{s}_j^\top,
\qquad
\mathbf{r}_{\text{feature}}
= \mathbf{P}\, \mathbf{z}.
\]

This residual captures the deviation between the predicted and observed bearing directions. This formulation avoids explicit scale and overparameterization, making it well-suited for optimization.

\paragraph*{Depth initialization and LRF-based priors}
Inverse-depth values are initialized using a hybrid strategy:
\begin{itemize}
    \item features near the optical center use depth from the LRF and receive a strong prior;
    \item features with sufficient parallax use triangulated depth;
    \item otherwise a nominal depth is assigned.
\end{itemize}
For LRF-constrained features, the depth prior takes the form
\[
r_{\text{prior}} = \rho - \rho_{\text{LRF}},
\]
directly coupling the 1D range measurement to the inverse depth of each feature.

While previous work has demonstrated the effectiveness of 1D range priors in low-speed flights, our experiments extend this validation to high-speed flight, where scale drift can accumulate rapidly. We observe that the LRF-based depth prior significantly enhances metric consistency in such scenarios, particularly during straight-line motion and hover phases. This aligns with theoretical insights from observability analysis \cite{xVio}, which shows that range constraints can help resolve scale in otherwise weakly observable motion regimes.

\paragraph*{Limitations}
While this method does have many advantages, there are weak points. A limitation of this approach is its reliance on accurate IMU data and reliable inverse depth estimates—both of which may degrade during high-agility maneuvers or hover.

\subsection{Barometer Compensation Model}
\label{sec:methods:barometercompensation}

To mitigate static pressure position error in barometric altitude estimates caused by high-speed flight, we employ a data-driven correction model based on a GRU neural network \cite{gru}. Recurrent architectures are naturally suited to this task, as SPPE is temporally correlated. We selected GRUs due to their ability to capture temporal dependencies while maintaining lower parameter complexity than Long Short-Term Memory (LSTM) networks \cite{LSTM}, and superior empirical performance over vanilla recurrent neural networks.

The input to the model has several components: roll, pitch, body-frame acceleration, body-frame velocity, and body-frame velocity squared. By expressing kinematics in the body frame and including roll and pitch, we enable the model to learn direction-dependent aerodynamic effects so long as they are represented in the training data. By not including yaw we ensure a yaw agnostic model that won't be influenced by training data collection methods. The GRU has a hidden size of 128 and three layers. The network has two output heads--one that outputs a scalar correction term and another that returns its log variance ($\log \sigma^2$). This allows us to dynamically scale the weight in the factor graph depending on the model's confidence.

The model was trained on a curated dataset designed to fully excite aerodynamic states affecting SPPE. We detail the data collection procedure below:
\begin{itemize}
    \item \textbf{Baseline Trajectories:} Straight-line north-south flights maintaining a constant heading for each out and back. The heading of the vehicle changes in 15° increments over a 180° frontal arc, repeating the same trajectory.
    \item \textbf{Speed Sweeps:} Baseline Trajectories repeated with velocity incremented in 5~m/s steps, including prolonged (30+~s) segments.
    \item \textbf{Bidirectional Repeats:} Each outbound trajectory was followed by a reversed inbound path.
    \item \textbf{Temporal Diversity:} Scenarios were repeated across different days and wind speeds to capture environmental variation.
\end{itemize}

Barometric and GPS altitudes were both low-pass filtered and aligned using the first 50 samples to remove static bias. All inputs were normalized. Ground truth for training was obtained from the GPS altitude. The model was trained with the Adam optimizer ($\text{lr}=10^{-3}$), using a sequence length of 4096 and a batch size of 64. We trained for 7 epochs using a focal Gaussian negative log-likelihood loss, defined as
\[
\mathcal{L}_{\text{focal}} =
\frac{1}{2N}
\sum_{i=1}^{N}
\Big(1 - e^{-\frac{(y_i - \hat{y}_i)^2}{2\sigma_i^2}}\Big)^{\!\gamma}
\left[
\log \sigma_i^2 +
\frac{(y_i - \hat{y}_i)^2}{\sigma_i^2}
\right],
\]
where $y_i$ is the ground-truth altitude correction, $\hat{y}_i$ is the predicted correction, $\sigma_i^2$ the predicted variance, and $\gamma$ is the focal exponent controlling the emphasis on high-residual (“hard”) samples. A warm-up schedule linearly increased $\gamma$ from $0$ to $1$ over the first three epochs, balancing stability and improved weighting of large standardized residuals.

At runtime, the GRU-corrected barometric measurement is fused with state estimates via the following residual:
\[
\mathbf{r}_{baro} = \mathbf{r}_3^\top (z_{\text{baro}} - b - z_{\text{est}}),
\]
where $\mathbf{r}_3$ is the third row of the rotation matrix (mapping body Z to world), $z_{\text{baro}}$ is the corrected barometric measurement, $b$ is a bias term relating world frame altitude with inertial frame altitude, and $z_{\text{est}}$ is the current estimated altitude. This model allows the barometric factor to dynamically adapt its influence based on learned uncertainty, enabling high-speed operation without compromising estimator stability. 

Because our GRU is trained on data from a single airframe, its correction is aircraft-specific and assumes ground speed $\approx$ airspeed; performance may degrade if those assumptions no longer hold. We did include several flights during high wind days (30+ mph) in the dataset and the compensation model was still quite effective. We empirically tested adding angular rates to the input, but found (in line with \cite{nn-baro}) that SPPE is dominated by velocity and acceleration so the current configuration yielded the best trade-off between accuracy and model simplicity.

\subsection{LRF Preprocessing}
\label{sec:methods:lrfpreprocessing}

The onboard LRF provides high-frequency ($\approx$10 Hz) altitude measurements relative to the terrain. While generally precise, its readings can exhibit transient spikes due to surface-dependent reflectivity, brief dropouts, or vehicle motion near high-contrast boundaries. To improve robustness and suppress these artifacts, we apply a gated Kalman filter based on a simple constant-velocity motion model. The filter estimates altitude and its rate of change, and uses statistical gating to reject measurements whose residuals exceed a Mahalanobis threshold, effectively removing both high-frequency noise and gross outliers.

The resulting altitude estimate is temporally smooth and reliable, making it suitable for backend optimization and feature validation. Its inverse is used both to initialize inverse depths for new features and to provide a prior for features observed near the center of the frame.





\subsection{Attitude Processing}
\label{sec:methods:magnetometerprocessing}

Rather than directly incorporating raw magnetometer measurements, our system leverages the fused attitude estimate provided by PX4. This estimate results from PX4’s onboard EKF, which tightly integrates magnetometer and IMU measurements to produce a drift-corrected orientation. In addition to providing a magnetically referenced yaw, the PX4 attitude prior helps maintain consistency between the onboard controller’s notion of orientation and the VIO estimate.

Although the PX4 attitude contains full 3D orientation information, we incorporate only its yaw component, since roll and pitch are already well constrained by the inertial preintegration and thermal-inertial odometry pipeline. The PX4-derived yaw prior is incorporated using the same residual structure typically used for magnetometer alignment:

\[
\mathbf{r}_{att} = \mathbf{v}_{\text{px4}} - ({}^{w}\mathbf{R}_{b})^\top \mathbf{v}_{\text{wld}},
\]
where $\mathbf{v}_{\text{px4}}$ is the horizontal-plane projection of the PX4 attitude's magnetic-north direction expressed in the body frame, $\mathbf{v}_{\text{wld}} = [1,0,0]^\top$ denotes magnetic north in the world frame, and ${}^{w}\mathbf{R}_{b} \in SO(3)$ is the estimated body-to-world rotation. Conceptually, this residual penalizes yaw deviation between the PX4’s magnetically referenced attitude and the VIO attitude estimate.

\subsection{Keyframe and State Structure}
\label{sec:methods:keyframestructure}

To manage computational complexity and support real-time operation, we adopt a sliding-window approach with a hybrid keyframe structure. A new keyframe is created every 20 frames. Each keyframe includes the full inertial navigation state: pose, velocity, orientation, and IMU biases $(\mathbf{b}_g, \mathbf{b}_a)$. The intermediate 20 normal frames include only the position, rotation, and velocity states and omit the bias terms. We refer to the keyframe before the intermediate frames as their parent keyframe.

This structure enables efficient modeling of IMU dynamics while limiting the number of high-dimensional states. Crucially, all IMU preintegration factors between frames share and reference the IMU biases from the most recent keyframe. These biases are assumed constant within each 20-frame group, and they serve as the basis for integrating inertial measurements across normal frames and to the next keyframe.

During backend optimization, preintegration residuals are formed between consecutive frames using the common bias estimate from the prior keyframe. This allows consistent inertial modeling without requiring bias estimation at every frame, reducing both memory and computation.

\subsection{Backend Optimization}
\label{sec:methods:backendoptimization}

The backend employs nonlinear least-squares optimization to jointly estimate the navigation state and feature inverse depths by minimizing a sum of weighted residuals from multiple sensor modalities. The overall optimization problem is formulated as:
\begin{equation}
\min_{\mathbf{x}} \sum_i \|\mathbf{r}_i(\mathbf{x})\|_{\mathbf{W}_i}^2,
\end{equation}
where $\mathbf{x}$ is the set of navigation states (position $\mathbf{p}$, inertial velocity $\mathbf{v}$, orientation $\mathbf{R}$, IMU biases $\mathbf{b}_g, \mathbf{b}_a$), and feature inverse depths $\rho$. The weight matrix $\mathbf{W}_i$ reflects the inverse covariance, balancing residual contributions according to measurement certainty.

To ensure computational efficiency without sacrificing accuracy, we adopt a selective relinearization strategy inspired by ICE-BA \cite{ICEBA}. Rather than relinearizing all variables at every iteration, only a subset whose change exceeds a predefined threshold are relinearized. This approach reduces the number of Jacobian and residual evaluations, thereby maintaining real-time feasibility even under high-rate estimation demands.

At each optimization iteration, the following steps are taken:
\begin{enumerate}
    \item Evaluate the change in the linearization point for each variable.
    \item Mark variables whose change exceeds a threshold.
    \item Apply the change, recompute Jacobians, and reevaluate residuals for the marked subset.
    \item Proceed with the LMPCG optimization.
\end{enumerate}

Optimization is performed using LMPCG \cite{pcg, ICEBA, BundleAdjustmentInTheLarge, ConjugateGradientBundleAdjustment, PushingTheEnvolopeOfModernBundleAdjustment}, which combines trust-region Levenberg–Marquardt (LM) damping with the efficiency of Preconditioned Conjugate Gradient (PCG). This combination is particularly beneficial for sparse and structured systems. The LM regularization introduces a damping term that stabilizes the solution by ensuring the Hessian remains positive definite, even when some parameters—such as feature inverse depths—are poorly constrained. The linearized system is given by:
\[
(\mathbf{J}^\top \mathbf{W} \mathbf{J} + \lambda \mathbf{I}) \Delta \mathbf{x} = -\mathbf{J}^\top \mathbf{W} \mathbf{r},
\]

where $\mathbf{J}$ is the Jacobian of residuals with respect to the state vector $\mathbf{x}$, $\mathbf{W}$ is a block-diagonal weight matrix, $\lambda$ is the LM damping parameter, and $\Delta \mathbf{x}$ is the proposed update vector.

We define the damped Hessian $\mathbf{H} = \mathbf{J}^T \mathbf{W} \mathbf{J} + \lambda \mathbf{I}$ and gradient vector $\mathbf{d} = -\mathbf{J}^T \mathbf{W} \mathbf{r}$, and partition them as:
\[
\mathbf{H} = \begin{bmatrix} \mathbf{B} & \mathbf{E} \\ \mathbf{E}^T & \mathbf{C} \end{bmatrix}, \quad \Delta \mathbf{x} = \begin{bmatrix} \Delta \mathbf{x}_B \\ \Delta \mathbf{x}_C \end{bmatrix}, \quad \mathbf{d} = \begin{bmatrix} \mathbf{d}_B \\ \mathbf{d}_C \end{bmatrix},
\]
where $\mathbf{B}$ and $\Delta \mathbf{x}_B$ correspond to the navigation state, $\mathbf{C}$ and $\Delta \mathbf{x}_C$ to the feature inverse depths, and $\mathbf{E}$ to the cross terms.

By applying the Schur complement, we eliminate the feature depths:
\begin{align*}
\mathbf{S} &= \mathbf{B} - \mathbf{E} \mathbf{C}^{-1} \mathbf{E}^\top, \\
\mathbf{v} &= \mathbf{d}_B - \mathbf{E} \mathbf{C}^{-1} \mathbf{d}_C, \\
\Delta \mathbf{x}_C &= \mathbf{C}^{-1} (\mathbf{d}_C - \mathbf{E}^\top \Delta \mathbf{x}_B),
\end{align*}
which is efficient since $\mathbf{C}$ is block-diagonal and easily invertible.

We then apply PCG to iteratively solve the reduced system $\mathbf{S} \Delta \mathbf{x}_B = \mathbf{v}$. PCG is preferred over QR factorization due to its superior efficiency in systems with tightly clustered eigenvalues. Additionally, as an iterative solver, PCG allows early termination after a fixed number of iterations, enabling a bounded runtime.

This approach preserves sparsity, improves robustness, and enables scalable optimization suitable for embedded, high-rate VIO systems. The resulting optimized state is fused with IMU measurements in the high-rate EKF. The EKF state is then forwarded to the flight controller at 30~Hz for closed-loop control.

\subsection{Marginalization Strategy}
\label{sec:methods:marginalization}

To maintain bounded computational complexity, we marginalize old states using a Schur-complement-based prior. Let $x_\text{oldest}$ denote the oldest keyframe and $x_\text{next}$ its successor. The procedure is:

\begin{enumerate}
    \item \textbf{Select states to remove.}  
    We collect all navigation and IMU-bias states whose parent keyframe is $x_\text{oldest}$ (including $x_\text{oldest}$ itself). We also gather all feature states whose root keyframe is $x_\text{oldest}$.

    \item \textbf{Copy feature states that continue beyond $x_\text{next}$.}
    If a feature has measurements both inside the marginalization window and also associated with $x_\text{next}$ or later, we split its representation. A cloned copy of the feature state (along with its prior, if present) is created and linked only to the factors that lie within the marginalization window; this cloned state will be marginalized. The original feature state is preserved, but instead of re-triangulating its inverse depth, we shift its root navigation state from $x_\text{oldest}$ to $x_\text{next}$ by inserting the known relative transform between these keyframes. All remaining factors are then re-associated with $x_\text{next}$.

    \item \textbf{Isolate the temporary marginalization block.}  
    Every factor involving any to-be-marginalized nav state or any cloned feature state is removed from the global containers and placed into a temporary set.  
    After this extraction, these factors no longer connect to any states that are being kept with the exception of $x_\text{next}$. Thus the global Hessian can be permuted into
    \[
      \mathbf{H}_\text{global} =
      \begin{bmatrix}
        \mathbf{H}_\text{marg} & 0 \\
        0 & \mathbf{H}_\text{keep}
      \end{bmatrix},
      \qquad
      \mathbf{r}_\text{global} =
      \begin{bmatrix}
        \mathbf{r}_\text{marg} \\
        \mathbf{r}_\text{keep}
      \end{bmatrix}.
    \]
    Marginalization therefore operates entirely on $\mathbf{H}_\text{marg}$.

    \item \textbf{Form the temporary Hessian and residual.}  
    We linearize all factors in the temporary container after applying any delayed state updates.  
    This produces a block-structured system
    \[
    \mathbf{H}_\text{marg} =
    \begin{bmatrix}
      \mathbf{H}_{11} & \mathbf{H}_{12} & \mathbf{H}_{13} \\
      \mathbf{H}_{12}^\top & \mathbf{H}_{22} & \mathbf{H}_{23} \\
      \mathbf{H}_{13}^\top & \mathbf{H}_{23}^\top & \mathbf{H}_{33}
    \end{bmatrix},
    \qquad
    \mathbf{r}_\text{marg} =
    \begin{bmatrix}
      \mathbf{r}_1 \\ \mathbf{r}_2 \\ \mathbf{r}_3
    \end{bmatrix},
    \]
    where:
    \begin{itemize}
        \item $\mathbf{H}_{11},\mathbf{r}_1$ contain contributions involving only the nav/IMU states being marginalized,
        \item $\mathbf{H}_{33},\mathbf{r}_3$ contain contributions involving only the cloned feature states,
        \item $\mathbf{H}_{22},\mathbf{r}_2$ contain contributions involving the state at $x_\text{next}$ (its nav state and IMU bias), and
        \item $\mathbf{H}_{12},\mathbf{H}_{13},\mathbf{H}_{23}$ are the cross-terms.
    \end{itemize}

    \item \textbf{Eliminate the cloned feature states.}  
    Since $\mathbf{H}_{33}$ is block-diagonal and therefore inexpensive to invert, we apply a Schur complement to remove the cloned feature states. Denoting the updated quantities with a tilde, 
    \begin{align*} 
        \tilde{\mathbf{r}}_1 &= \mathbf{r}_1 - \mathbf{H}_{13}\mathbf{H}_{33}^{-1}\mathbf{r}_3, \\ \tilde{\mathbf{r}}_2 &= \mathbf{r}_2 - \mathbf{H}_{23}\mathbf{H}_{33}^{-1}\mathbf{r}_3, \\ \tilde{\mathbf{H}}_{11} &= \mathbf{H}_{11} - \mathbf{H}_{13}\mathbf{H}_{33}^{-1}\mathbf{H}_{13}^\top, \\ \tilde{\mathbf{H}}_{22} &= \mathbf{H}_{22} - \mathbf{H}_{23}\mathbf{H}_{33}^{-1}\mathbf{H}_{23}^\top, \\ \tilde{\mathbf{H}}_{12} &= \mathbf{H}_{12} - \mathbf{H}_{13}\mathbf{H}_{33}^{-1}\mathbf{H}_{23}^\top.
    \end{align*}

    \item \textbf{Eliminate the nav/IMU states between keyframes.}  
    We next remove the remaining nav states associated with $x_\text{oldest}$ using the same method:
    \begin{align*}
      \bar{\mathbf{H}}_{22}
      &= \tilde{\mathbf{H}}_{22}
        - \tilde{\mathbf{H}}_{12}^\top \tilde{\mathbf{H}}_{11}^{-1} \tilde{\mathbf{H}}_{12},
      \\
      \bar{\mathbf{r}}_{2}
      &= \tilde{\mathbf{r}}_{2}
        - \tilde{\mathbf{H}}_{12}^\top \tilde{\mathbf{H}}_{11}^{-1} \tilde{\mathbf{r}}_{1}.
    \end{align*}
    The pair $(\bar{\mathbf{H}}_{22}, \bar{\mathbf{r}}_2)$ captures all information from the removed nav, IMU, and feature states as a prior on the keyframe $x_\text{next}$.

    \item \textbf{Regularization and prior factor construction.}  
    We symmetrize and clip the singular values of $\bar{\mathbf{H}}_{22}$ for numerical stability, then compute the implied prior mean
    \[
      \boldsymbol{\mu}_\text{prior} = \bar{\mathbf{H}}_{22}^{-1}\bar{\mathbf{r}}_2.
    \]
    A Nav+IMU-bias prior factor is constructed using $(\bar{\mathbf{H}}_{22},\boldsymbol{\mu}_\text{prior})$ and attached to $x_\text{next}$.  
    The marginalized states and all of their factors are then discarded.
\end{enumerate}

\section{Results}
\label{sec:results}

\begin{figure*}[h]
  \centering
  \includegraphics[width=0.9\linewidth]{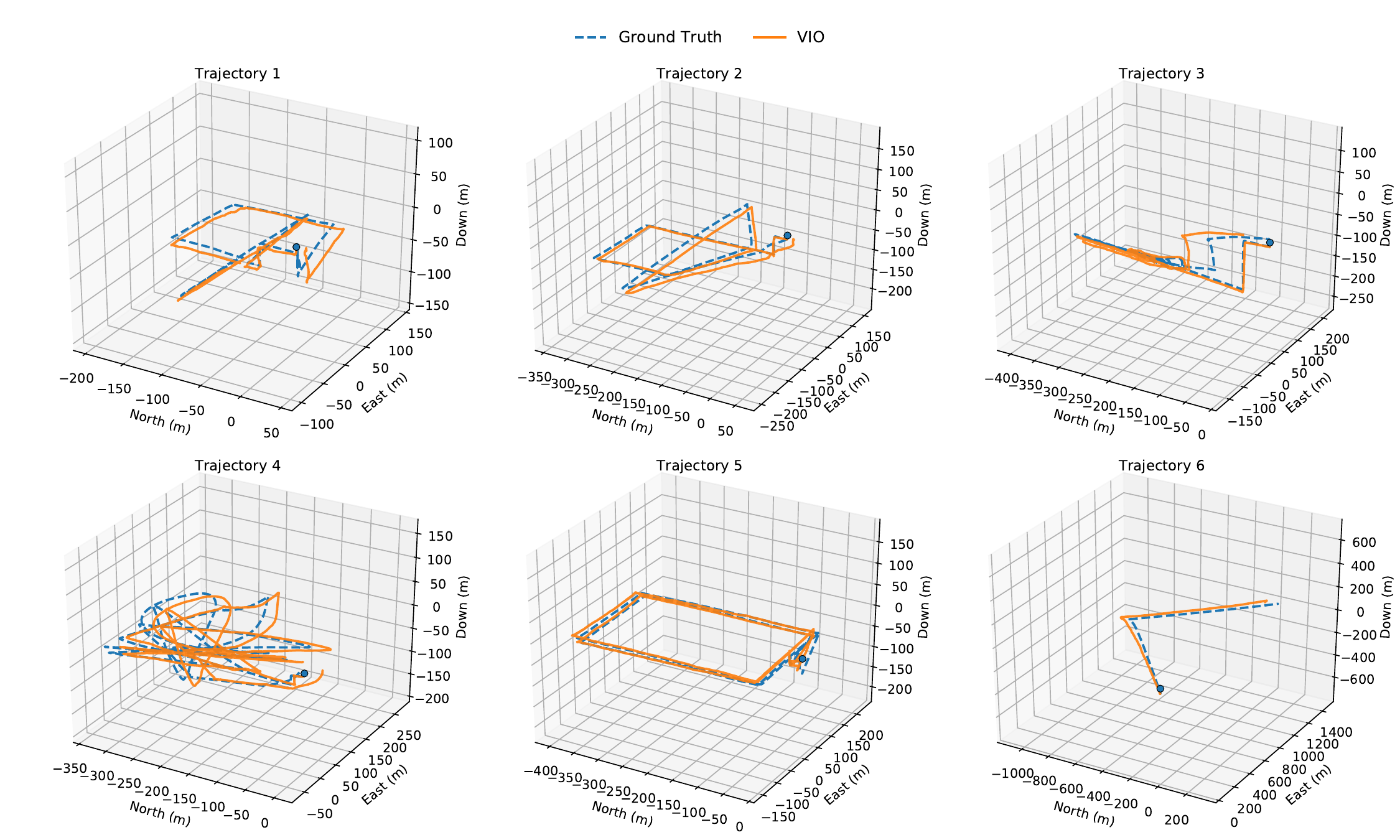}
    \caption{Example trajectories from flight experiments. The ground truth reference was obtained using onboard GPS measurements.}
  \label{fig:trajectories}
\end{figure*}

In this section, we present experimental validation of the methods presented with an emphasis on demonstrating our stated contributions. We systematically evaluate the impact of each component on system performance through dedicated experiments. First, we demonstrate the effectiveness and robustness of our feature tracking pipeline by comparing it to a baseline. Second, we show the improvements made by our SPPE compensator to the estimated trajectory and compare it to existing methods. Next, we show the improvement from prefiltering the high-rate IMU data before downsampling. Lastly, we examine closed-loop flight performance both in estimator drift and timing characteristics. To enable direct comparison, for flights other than the closed-loop experiment, logged data was replayed on a laptop with individual pipeline components disabled. Ground truth trajectories are derived from onboard GPS. A sample of the types of trajectories flown can be seen in Figure \ref{fig:trajectories}. 

To quantify estimator performance, we report both the root-mean-square error (RMSE) and the drift rate for each trajectory. RMSE is computed by first time-aligning the estimated and GPS-derived trajectories. We interpolate the GPS measurements to the estimator timestamps and then evaluate the Euclidean position error at each matched sample. The reported RMSE is the square root of the mean squared position error over the full flight. Drift is measured as the final position error divided by the total distance traveled, expressed as a percentage, providing a scale-normalized measure of long-term consistency. Together, these metrics characterize both the instantaneous accuracy and long-horizon stability of the estimator.

\subsection{Experiment 1: Thermal-Optimized Feature Pipeline}
\label{sec:results:featurefiltering}

We evaluate our thermal-optimized feature tracking pipeline against a baseline composed of Gaussian filtering, CLAHE \cite{CLAHE}, FAST feature detection \cite{fast-features}, Lucas–Kanade optical flow \cite{lkt}, and essential-matrix–based outlier rejection. Our enhancements—gradient-based filtering, ORB descriptor \cite{orb} filtering, and triangulation-based filtering—are introduced incrementally to quantify their individual and combined effects. To ensure diversity in testing, five representative flight logs were selected, with peak speeds ranging from 20~m/s to 30~m/s. Two of these logs were chosen specifically for their elevated fixed-pattern noise (FPN) levels that exceed typical conditions. Table~\ref{tab:feat_ablation} summarizes the resulting RMSE mean and standard deviation across all logs. As shown, each added filtering stage either reduces overall error magnitude or decreases the variability of that error, with the full filtering configuration consistently delivering the best performance across all datasets. While the table reports cumulative improvements, we note that each filtering step also improves performance when applied in isolation, though we omit those results for brevity.

\begin{table}[h]
  \caption{Experiment 1: Feature pipeline ablation}
  \label{tab:feat_ablation}
  \centering
  \begin{tabular}{lcc}
    \toprule
    {\bf Method}                                    & {\bf RMSE Mean (m)} & {\bf RMSE STD (m)} \\
    \midrule
    Baseline                                    & 19.35          & 7.03             \\
    + Gradient filtering                           & 16.63          & 7.63             \\
    + ORB descriptor filtering                     & 12.43          & 5.23            \\
    + Triangulation filtering                      & 12.83          & 3.65             \\
    \bottomrule
  \end{tabular}
\end{table}

\subsection{Experiment 2: Barometric Compensation Efficacy}
\label{sec:results:barocompensation}

\begin{figure*}[!ht]
  \centering
  \includegraphics[width=0.95\linewidth]{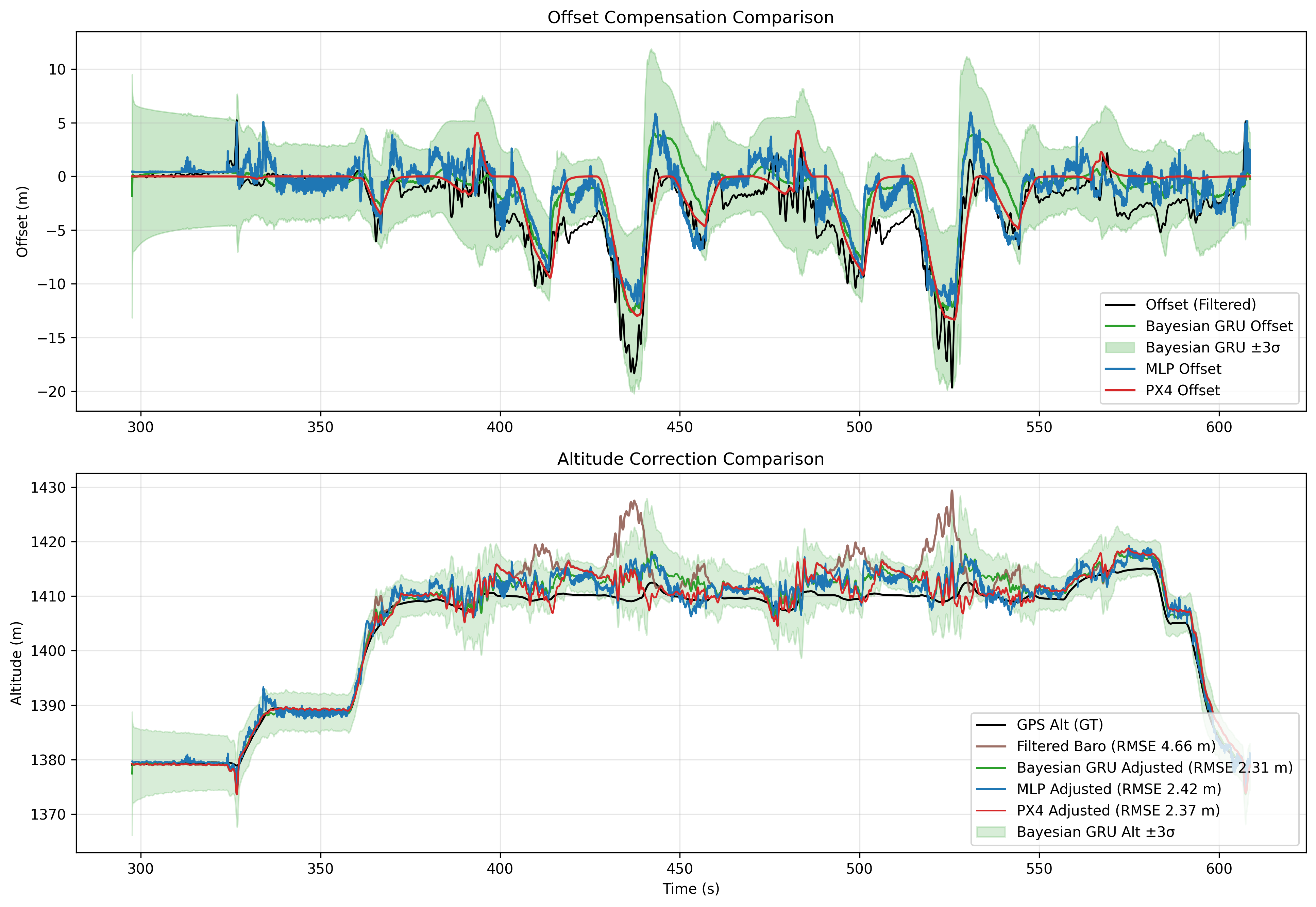}
  \caption{Barometric error vs.\ time for filtered, polynomial, MLP, and GRU corrections.}
  \label{fig:baro_error}
\end{figure*}

To evaluate our GRU-based barometric compensator, we compare it against conventional polynomial fitting (like PX4 \cite{px4} and Ardupilot \cite{ardupilot}) and an MLP-based method using a withheld test dataset. Figure \ref{fig:baro_error} shows the GPS altitude and the low-pass filtered barometer altitude compared against the three compensation methods for a sample trajectory. Additionally, we compare the methods' effects on the estimated trajectory using five example flights. As shown in Fig.~\ref{fig:baro_error}, the GRU correction most closely tracks GPS altitude on a representative high-speed trajectory. On this sequence, RMSE is reduced from 4.66~m (filtered baro) to 2.19~m with the GRU, a 2.47~m improvement. Further, the GRU-based method not only performs better based on the mean, but the three sigma bound on uncertainty nearly encapsulates the error and further enhances performance. Similar improvements are observed across the full validation set (Table~\ref{tab:baro_rmse}).

Table \ref{tab:baro_rmse} shows the RMSE for the withheld test set as well as the average Z RMSE, average trajectory RMSE, average end point error (EPE), and average trajectory drift rate for the five high speed flights under consideration. In each case, our proposed GRU-based compensator outperforms the other methods.

\begin{table}[h]
  \caption{Barometric Compensation Method Comparison}
  \label{tab:baro_rmse}
  \centering
  \setlength{\tabcolsep}{3pt}
  \begin{tabular}{lcccc}
    \toprule
    {\bf Method}        & {\bf Test RMSE} & {\bf Z RMSE} & {\bf Trajectory RMSE} & {\bf Drift (\%)} \\
    \midrule
    Filtered baro       & 3.35         & 4.07         & 38.93          & 3.20         \\
    Poly fit            & 2.60         & 2.93         & 29.14          & 2.11         \\
    MLP                 & 2.59         & 3.58         & 31.99          & 2.12         \\
    \textbf{GRU (ours)} & \bf 2.00     & \bf 2.82     & \bf 25.03      &  \bf 1.27    \\
    \bottomrule
  \end{tabular}
\end{table}

\begin{figure} [t]
    \centering
    \includegraphics[width=0.9\linewidth]{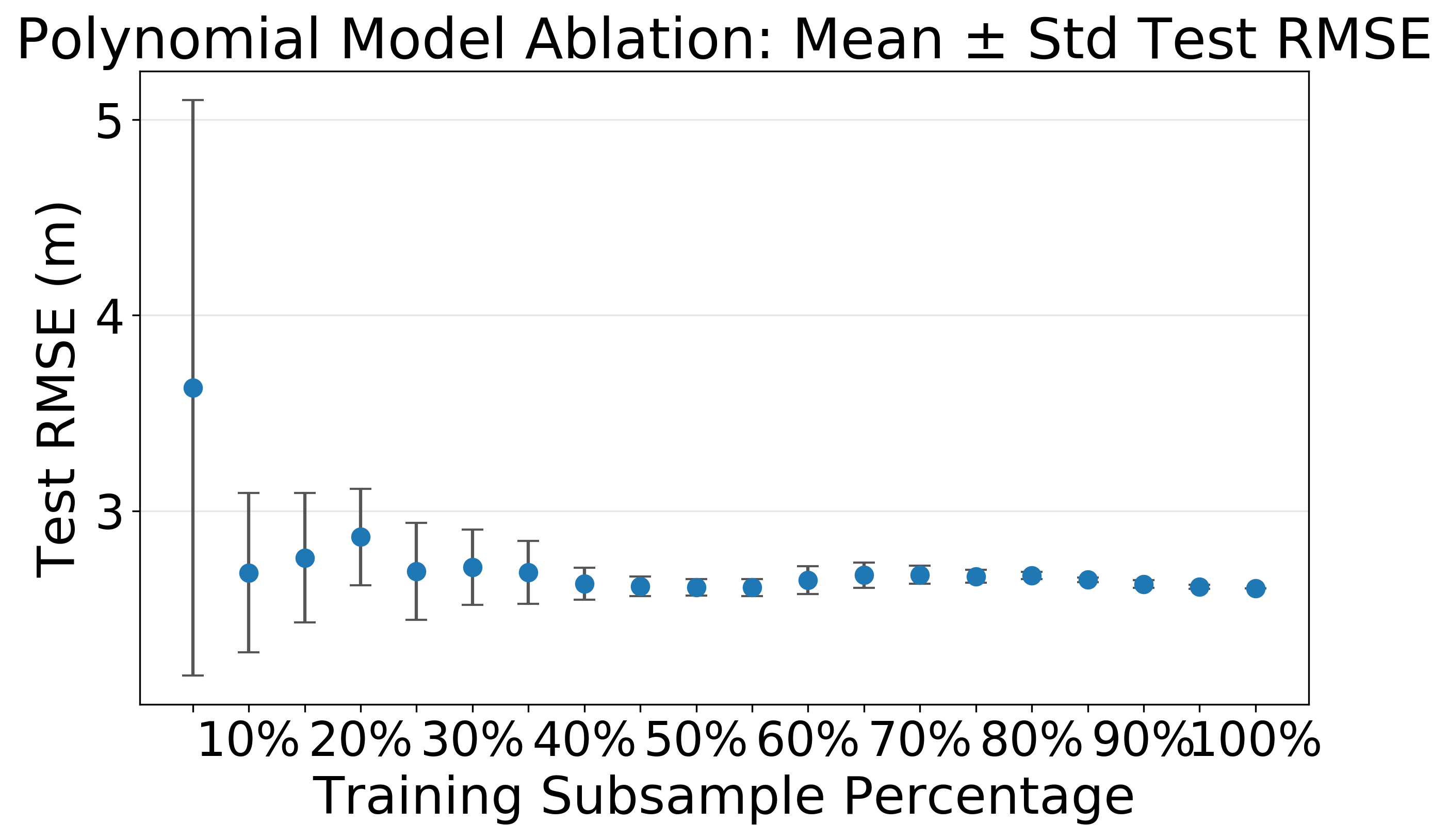}
    \caption{Mean and STD of test error for polynomial models trained with random sub samples of training data}
    \label{fig:baro_training}
\end{figure}

One downside of the deep learning based methods is the large amount of data required to train an effective model. Because the polynomial fit method has fewer parameters, it's intuitive to assume that it would require less data while still achieving the desired performance gain. In order to test this intuition we developed a Monte Carlo training scenario. For each training test a percentage of the training data is randomly selected as a block from the total training set. The model is then trained with the subsampled data and tested against the withheld validation set. For each percentage level, this process was repeated 30 times and the mean and standard deviation of the distribution computed. As can be seen in Fig. ~\ref{fig:baro_training}, more than 45\% of the data collected is necessary before the results become reliable. This indicates that the perceived data-efficiency advantage of the polynomial method is, in practice, much less than assumed. This is likely because sufficient diversity in orientation is needed to build a representative model of the aerodynamics. This experiment was repeated for our GRU-based method with similar results. 

Although it has more tunable parameters than the polynomial fit, our GRU-based method remains lightweight and real-time compatible. Our implementation uses only 252,546 parameters. It incurs negligible runtime overhead and provides superior accuracy once trained.

\subsection{Experiment 3: Impact of high‐rate IMU prefiltering}
\label{sec:results:imuprefiltering}
We compare estimator performance using raw IMU measurements at 120 Hz versus our proposed IIR-filtered and downsampled method. The results in Table ~\ref{tab:imu_ablation} show the EPE, drift rate, and RMSE for an illustrative 4 km trajectory with many abrupt changes in direction. A plot of the trajectory in question, along with our estimated trajectory, can be seen as Trajectory 4 in Figure \ref{fig:trajectories}. These results demonstrate that our prefiltering and downsampling method provides better performance than the non-filtering method in aggressive flight regimes. It also avoids saturating the limited communication bandwidth between processors, improving reliability.

\begin{table}[h]
  \caption{IMU processing comparison}
  \label{tab:imu_ablation}
  \centering
  \begin{tabular}{lccc}
    \toprule
    {\bf Method}                          & {\bf EPE (m)} & {\bf Drift (\%)} & {\bf RMSE (m)} \\
    \midrule
    120~Hz raw preintegration            & 150.07          & 3.60       & 77.68            \\
    \textbf{1200 Hz + downsampling}     & \bf 31.24      & \bf 0.75     & \bf 24.96         \\
    \bottomrule
  \end{tabular}
\end{table}

\subsection{Experiment 4: Closed‐loop flight and latency}
\label{sec:results:closedloop}

We validate real-time closed-loop flight performance of the proposed estimator on an NVIDIA Jetson Xavier NX using a single box-pattern trajectory flown at 30~m/s. The onboard estimator provided pose updates at 30~Hz to the flight controller for fully closed-loop operation. Table~\ref{tab:latencies} summarizes the average execution latency of each major pipeline component. The full estimation loop maintained an average cycle time of 42~ms, enabling closed-loop flight even during aggressive maneuvers.

Table~\ref{tab:closed_loop_error} reports the final trajectory accuracy relative to GPS ground truth. Over the 3.1~km box pattern, the estimator achieved an endpoint position error of 31.5~m, corresponding to a drift rate of 1.0\%. These results demonstrate that the system is capable of delivering state estimates for high-speed, GPS-denied flight while operating fully in real time on embedded hardware.

\begin{table}[h]
  \caption{Closed‐loop latencies (Xavier NX)}
  \label{tab:latencies}
  \centering
  \begin{tabular}{lcc}
    \toprule
    {\bf Component}                 & {\bf Mean (ms)} & {\bf Std} \\
    \midrule
    Image Preprocessing            & 11.63             & 2.91               \\
    Feature Detection              & 1.74             & 1.08              \\
    Feature Tracking               & 4.66             & 1.60              \\
    Graph Optimization + Marginalization             & 23.9             & 23.4              \\
    \midrule
    \textbf{Total loop time}       & \bf 41.93        & \bf 28.99          \\
    \bottomrule
  \end{tabular}
\end{table}

\begin{table}[h]
  \caption{Closed Loop Error}
  \label{tab:closed_loop_error}
  \centering
  \begin{tabular}{ccc}
    \toprule
    {\bf EPE(m)}                          & {\bf Drift (\%)} & {\bf RMSE (m)} \\
    \midrule
    31.51            & 1.00          & 19.23             \\
    \bottomrule
  \end{tabular}
\end{table}



\section{Conclusion}
\label{sec:conclusion}

In this paper, we presented a monocular thermal-inertial odometry system specifically designed for high-speed flight in visually degraded, GPS-denied environments. Our comprehensive approach addressed key challenges unique to thermal imagery and high velocities, integrating a robust thermal-optimized feature tracking pipeline, high-rate IMU prefiltering, and a novel GRU-based barometric compensation model.

Through rigorous experimental evaluations, we demonstrated significant performance improvements from each component. Our optimized feature tracking pipeline markedly reduced estimation drift and endpoint errors compared to conventional methods. The proposed GRU-based barometric compensator consistently outperformed traditional polynomial and MLP-based corrections, effectively capturing temporal dependencies and enhancing altitude estimation accuracy. Additionally, our IMU prefiltering strategy significantly improved estimator stability during aggressive maneuvering and enabled efficient bandwidth utilization without compromising performance.

Collectively, these advancements enabled robust, real-time thermal-inertial odometry at flight velocities of 30 m/s. The system maintained low drift (\(<\)1.3\%) and small average RMSE (\(<\)26 m), validating its suitability for demanding aerial robotics applications. 

Future work includes extending our system to multi-sensor fusion approaches involving EO/IR combinations, leveraging IMU data to better filter feature tracks, and improved thermal camera to IMU calibration.

\ifCLASSOPTIONcaptionsoff
  \newpage
\fi



%
\balance
\newpage

\bibliographystyle{IEEEtran}
\bibliography{bib/bibliography}  

\end{document}